\documentclass{article}

    \PassOptionsToPackage{numbers, compress}{natbib}

    \usepackage[final]{neurips_2021}

\usepackage[utf8]{inputenc} %
\usepackage[T1]{fontenc}    %
\usepackage{hyperref}       %
\usepackage{url}            %
\usepackage{booktabs}       %
\usepackage{amsfonts}       %
\usepackage{nicefrac}       %
\usepackage{microtype}      %
\usepackage{url}
\usepackage{graphicx}
\usepackage{xcolor}
\usepackage{subcaption}
\usepackage{array}
\usepackage{booktabs}
\usepackage{stmaryrd}
\usepackage{amsmath}
\usepackage{tabularx}
\usepackage{verbatim}
\usepackage{tikz}
\usetikzlibrary{calc,positioning}
\usetikzlibrary{shapes.geometric}   %
\usepackage{rotating}               %
\usetikzlibrary{positioning,arrows.meta,calc}
\usepackage{adjustbox}
\title{InvGAN: Invertible GANs}

\author{%
  Partha Ghosh\thanks{This work was completed while the author was an intern with Amazon} $\, ^\dagger$ 
   \And
   Dominik Zietlow$^{* \dagger}$
   \And
   Michael J. Black$^\ddag$
   \AND
   Larry S. Davis $^\ddag$
   \And
   Xiaochen Hu $^\ddag$ \\
\AND
\texttt{$\dagger$ MPI for Intelligent Systems, T{\"u}bingen \quad \qquad $\ddag$ Amazon.com, Inc.}\\
\texttt{\{pghosh, dzietlow\}@tue.mpg.de \quad \{mjblack, lrrydav, sonnyh\}@amazon.com}\\
}

\newcommand{\qheading}[1]{\noindent\textbf{#1}}

\newcommand{\norm}[1]{\left\lVert#1\right\rVert}

\newcommand{\eg}{e.g.\ }
\newcommand{\ie}{i.e.\ }
\newcommand\doubleplus{\ensuremath{\mathbin{+\mkern-10mu+}}}

\newcommand{\sZ}{\mathbb{Z}}
\newcommand{\sX}{\mathbb{X}}
\newcommand{\sW}{\mathbb{W}}

\begin{document}

\maketitle

\begin{abstract}
Generation of photo-realistic images, semantic editing and representation learning are a few of many potential applications of high resolution generative models. Recent progress in GANs have established them as an excellent choice for such tasks.
However, since they do not provide an inference model, image editing or downstream tasks such as classification can not be done on real images using the GAN latent space. 
Despite numerous efforts to train an inference model or design an iterative method to invert a pre-trained generator, previous methods are dataset (\eg human face images) and architecture (\eg StyleGAN) specific.  These methods are nontrivial to extend to novel datasets or architectures. We propose a general framework that is agnostic to architecture and datasets. Our key insight is that, by training the inference and the generative model together, we allow them to adapt to each other and to converge to a better quality model. Our \textbf{InvGAN}, short for Invertible GAN, successfully embeds real images to the latent space of a high quality generative model. This allows us to perform image inpainting, merging, interpolation and online data augmentation.
We demonstrate this with extensive qualitative and quantitative experiments.
\end{abstract}
\section{Introduction}

The ability to generate photo-realistic images of objects such as human faces or fully clothed bodies has wide applications in computer graphics and computer vision. 
Traditional computer graphics, based on physical simulation,
often fails to produce photo-realistic images of objects with
complicated geometry and material properties.
In contrast, modern data-driven methods, such as deep learning based generative models show great promise for realistic image synthesis~\cite{karras2019style, karras2019analyzing}. 
Among the four major categories of generative models, generative adversarial networks (GANs), variational auto-encoders (VAEs), normalizing flow networks and autoregressive models, GANs deliver images with the best visual quality. Although recent efforts in VAEs \cite{child2021very, razavi2019generating} have tremendously improved their generation quality, they still use larger latent space dimensions and deliver lower quality images. Autoregressive models are very slow to sample from and do not provide a latent representation for the trained data. Finally, flow-based methods do not perform dimensionality reduction and hence produce large models and latent representations. On the other hand, GANs do not provide a mechanism to embed real images into the latent space.
This limits them as a tool for image editing and manipulation.
Specifically, while several methods exist, there is no  method that trains a GAN so that it can be efficiently and effectively inverted.
To that end, we propose {\em InvGAN}, an invertible GAN 
with an inference module that can embed real images into the latent space.
InvGAN has wide range of applications.

\qheading{Representation Learning.}
GANs learn a latent representation of the training data.
This representation has been shown to be well structured \cite{DBLP:journals/corr/abs-1809-11096, karras2019style, karras2019analyzing},
allowing GANs to be employed for a variety of downstream tasks (\eg classification, regression and other supervised tasks) \cite{9304879,ramaswamy2020fair}.
We extend the GAN framework to include an inference model that embeds real images into the latent space. 
InvGAN addresses this problem and
can be used to support representation learning \cite{chen2020generative,locatello2019challenging}, data augmentation \cite{DBLP:journals/corr/abs-1809-11096,DBLP:journals/corr/abs-1904-09135} and algorithmic fairness \cite{balakrishnan2020towards,sattigeri2018fairness,sharmanska2020contrastive}. 
Previous methods of inversion rely on computationally expensive optimization of inversion processes, limiting their scope to offline applications, \eg data augmentation has to happen before training starts.
Efficient, photo-realistic, semantically consistent, and model based inversion is the key to online and adaptive use-cases.

\qheading{Conditional Image Editing.}
Recent work shows that even unsupervised GAN training isolates several desirable generative characteristics~\cite{nguyen2019hologan, voynov2020unsupervised}. 
Prominent examples are correspondences between latent space directions and \eg hair style, skin tone and other visual characteristics.
Recent works provide empirical evidence suggesting that one can find paths in the latent space (albeit non-linear) that allow for editing individual semantic aspects. GANs therefor have the potential to become a high-quality graphics editing tool \cite{GIF2020, tewari2020stylerig}. 
However without a reliable mechanism for projecting real images into the latent space of the generative model, editing of real data is impossible. 
InvGAN take a step towards addressing this problem.

\section{Related work}
\label{related_work}

The GAN inversion task has been addressed in two primary ways (1) using an inversion model (often a deep neural network), (2) embedding real images into the latent space of a trained generator using an iterative optimization based method, typically initialized with a deep model. 
 
\qheading{Optimization based:}
iGAN~\cite{zhu2016generative} optimizes for a latent code while minimizing the distance between a generated image and a source image. 
To ensure uniqueness of the preimage of a GAN-generated data point, Zachary et al.~\cite{lipton2017precise} employ stochastic clipping. 
As the complexity of the GAN generators increases, an inversion process based on gradient descent and pixel space $\text{MSE}$ is insufficient. 
Addressing this, Rameen et al.~specifically target StyleGAN generators and optimize for perceptual loss \cite{DBLP:journals/corr/abs-1904-03189, DBLP:journals/corr/abs-1911-11544}. However, they invert into the $W+$ space, the so called extended $w$ space of StyleGAN. This results in high dimensional latent codes and consequently prolongs inversion time. 
This can also produce out-of-distribution latent representations which makes them unsuitable for downstream tasks. 
Contrary to these drawbacks InvGAN offers fast inference embedding in the non-extended latent space.%

\qheading{Model based:} %
BiGAN~\cite{donahue2016adversarial} and ALI~\cite{dumoulin2016adversarially} invert the generator of a GAN during the training process by learning the joint distribution of the latent vector and the data in a completely adversarial setting. 
However, the quality is limited, partially because of the choice of DCGAN~\cite{radford2015unsupervised} and partially because of the significant dimensionality and distribution diversity between the latent variable and the data domain \cite{DBLP:journals/corr/abs-1907-02544}. %
More recent models target the StyleGAN architecture \cite{DBLP:journals/corr/abs-2008-00951,wei2021simple, DBLP:journals/corr/abs-1906-08090} and achieve impressive results. Most leverage StyleGAN peculiarities, i.e. they invert in the $W+$ space -- so adaptation to other GAN backbones is non-trivial. 
Adversarial latent auto-encoders \cite{DBLP:journals/corr/abs-2004-04467}, are closest to our current work. 
Our model and adversarial autoencoders can be made equivalent with a few alterations to the architecture and to the optimization objective. We discuss this more in detail in Section \ref{sub_sec:objective}. Our method in contrast to previous works discussed in this section, neither uses any data set specific loss nor does it depend upon any specific network architecture.

\qheading{Hybrid optimization and regression based:}
Guan et al.~\cite{DBLP:journals/corr/abs-2007-01758} train a regressor that is used to initialize an optimization-based method to refine the regressor's guess.
However, to achieve good results, this method uses an identity loss to guide the refinement procedure making it specific to human face datasets. 
Zhu et al.~\cite{zhu2020indomain} modify the general hybrid approach with an additional criterion such that the recovered latent code must belong to the semantically meaningful region learned by the generator.
It is thereby assumed that the real image can be reconstructed more faithfully in the immediate neighbourhood of this initial guess.
Yuval et al.~\cite{alaluf2021restyle} replace  gradient-based optimization with an
iterative encoder that encodes the current generation and target image to the estimated latent code difference. They empirically show that this iterative process converges and the recovered image improves over iterations. However, this method requires multiple forward passes in order to achieve a suitable latent code. In contrast to the work above, the inference module obtained by our method infers the latent code in one shot. Hence it is much faster and does not run the risk of finding a non-meaningful latent code. %

The inversion mechanisms presented so far do not directly influence the generative process. In most of the cases, they are conducted on a pre-trained frozen generator. Although in the case of ALI~\cite{dumoulin2016adversarially} and BiGAN~\cite{donahue2016adversarial}, the inference model loosely interacts with the generative model at training time.
However, the interaction is only indirect; \ie through the discriminator. In our work we tightly couple the inference module with the generative module, resulting in better reconstruction quality.

\qheading{Joint training of generator and inference model:} We postulate that jointly training an inference module will help regularize GAN generators towards invertability. This is inspired by the difficulty of inverting a pre-trained high-performance GAN. For instance Bau et al.~\cite{bau2020semantic} invert PGAN~\cite{karras2017progressive}, but for best results a two-stage mechanism is needed. Similarly Image2StyleGAN~\cite{abdal2019image2stylegan} projects real images into the extend $w^+$ space of StyleGAN, whereas, arguably, all the generated images can be generated from the more compact $z$ or $w$ space.
This is further evident from Wulff et al.~\cite{wulff2020improving} who find an intermediate latent space in StyleGAN that is more Gaussian than the assumed prior. However, they too use an optimization-based method and, hence, it is computationally expensive and at the same time specific to both the StyleGAN backend and the specific data set.
Finally we refer the readers to `GAN Inversion: A Survey' \cite{DBLP:journals/corr/abs-2101-05278} for a comprehensive review of relate work.
\section{Method}
\label{method}
\qheading{Goal}: Our goal is to learn an inversion module alongside the generator during GAN training. Specifically we find a generator $G: \sW \rightarrow \sX$ and an inference model $D: \sX \rightarrow \sW$ such that $x \approx G(D(x \sim \sX))$. Where $\sX$ denotes the data domain and $\sW$ denotes the latent space. By doing so we (1) unlock semantic editability of real images, (2) allow semi-supervised learning, (3) encourage latent space smoothness, within the GAN framework. 

\subsection{Architecture}
We demonstrate InvGAN using DC-GAN, BigGAN and StyleGAN as the underlying architectures. Figure \ref{figure:invgan_model} represents the schematic of our model. We follow the traditional alternative generator-discriminator training mechanism. The generative part consists of three steps $z \sim \mathcal{N}(0, I); w = M(z); x = G(w)$, where $M$ is a mapping network,  $G$ is the generator, $D$ is the discriminator, and $\mathcal{N}(0, I)$ is the standard normal distribution. In the generator, we use the standard $8$-layer mapping network with StyleGAN and add a $2$-layer mapping network to BigGAN and DC-GAN. The discriminator, besides outputing real/fake score, also outputs inferred $w$ parameter. From here on we use $\tilde{w}, c = D(x)$ to denote the inferred latent code ($\tilde{w}$) using the discriminator $D$ and $c$ to denote the real-fake classification decision for the sample $x\in \sX$. Wherever obvious we simply use $D(x)$ to refer to $c$, the discrimination decision only.

\subsection{Objective}
\label{sub_sec:objective}
\qheading{GAN Objective:} The min-max game between the discriminator network and the generator network of vanilla GAN training is described as
\begin{equation}
    \min_{G, M}\max_{D}\mathcal{L}_\text{GAN} = \min_{G, M}\max_{D}\left[\mathbb{E}_{x \in \sX}[\log D(x)] + \mathbb{E}_{z\in \sZ}[\log(1-D(G(M(z))))]\right].
\label{eq:gan_obj}
\end{equation}
 A naive attempt at an approximately invertible GAN would perform $\min_{G}\max_{D}L_\text{GAN} + \min_D\norm{w - \tilde{w}}_p$,  where $\norm{\bullet}_p$ denotes an $L_p$ norm. This loss function can be interpreted as an optimal transport cost. We discuss this in more detail at the end of this section.
 However, this arrangement, coined the "naive model", does not yield satisfactory results, cf.~Section \ref{sec:ablation_resue_real_z}. This can be attributed to two factors: (1) $w$ corresponding to real images are never seen by the generator; (2) no training signal is provided to the discriminator for inferring the latent code corresponding to real images ($w_R$); (3) the distribution of $w_R$ might differ from prior distribution of $w$. We address each of these concerns with a specific design choice. Our naive model corresponds to the adversarial autoencoders \cite{DBLP:journals/corr/abs-2004-04467} if the real-fake decision is derived from a common latent representation.
However, this forces the encoding of real and generated images to be linearly separable and contributes to degraded inference performance. 

\qheading{Minimizing latent space extrapolation:} Since, in the naive version, neither the generator nor the discriminator gets trained with $w_R$, it relies completely upon its extrapolation characteristics. In order to reduce the distribution mismatch for the generator we draw half the mini batch of latent codes from the prior and the other half consists of $w_R$; i.e. $w_\text{total} = w  \doubleplus w_R,\ w \sim P(W)$ where $\doubleplus$ denotes a batch concatenation operation. By $w \sim P(W)$ we denote the two stage process given by the following $w = M(z\sim P(Z))$.

\qheading{Pixel space reconstruction loss:} Since latent codes for real images are not given, the discriminator cannot be trained directly. However, we recover a self-supervised training signal by allowing the gradients from the generator to flow into the discriminator. Intuitively, the discriminator tries to infer latent codes from real images that help the generator reproduce that image. As shown in Section \ref{sec:ablation_resue_real_z}, this helps improving real image inversion tremendously. We enforce further consistency by imposing an image domain reconstruction loss between input and reconstructed real images. However, designing a meaningful distance function for images is a non-trivial task. Ideally we would like a feature extractor function $f$ that extracts low- and high-level features from the image such that two images can be compared meaningfully. Given such a function a reconstruction loss can be constructed as %
\begin{equation}
\mathcal{L}_\text{fm} = \norm{\mathbb{E}_{x \in \sX}(f(x) - f(G(w \sim P(W|x)))}_p
\label{eq:feature_matching}
\end{equation}

\begin{figure*}
    \centering
    \includegraphics[width=\linewidth]{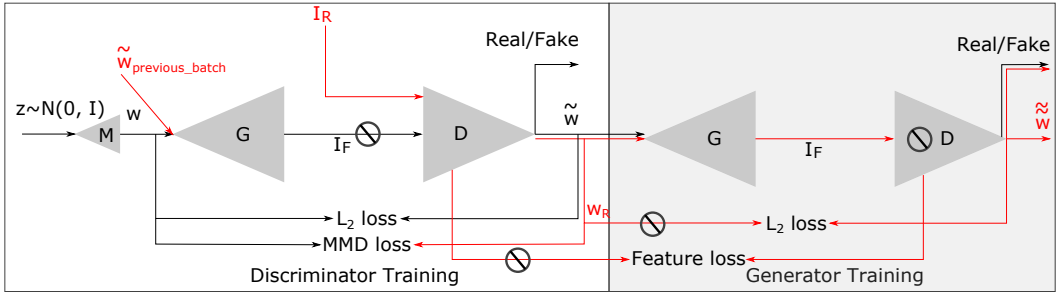}
    \caption{We train InvGAN following a regular GAN. We use a second output head in the discriminator besides the real fake decision head, to infer the latent-code $z$ of a given image. Here $\varobslash$ denotes no gradient propagation during back propagation step. It also denotes `no training' when it is placed on a model. We use red color to show data flow corresponding to real images.}
    \label{figure:invgan_model}
\end{figure*}

A common practice in the literature is to use a pre-trained VGG \cite{DBLP:journals/corr/JohnsonAL16, zhang2018perceptual} network as a feature extractor $f$. 
However it is well known that deep neural networks are susceptible to adversarial perturbations. Given this weakness, optimizing for perceptual loss is error prone. Hence a combination of a pixel-domain $L_2$ and feature-space loss is typically used. This often results in degraded quality.
Consequently, we take the discriminator itself as the feature extractor function $f$. 
Due to the min-max setting of GAN training, we are guaranteed to avoid the perils of adversarial and fooling samples.
The feature loss is shown in the second half of Figure \ref{figure:invgan_model}. 
Although this resembles the feature matching described by Salimans et al.~\cite{DBLP:journals/corr/SalimansGZCRC16}, it has a crucial difference. As seen in Equation \ref{eq:feature_matching} the latent code fed into the generator is drawn from the conditional distribution $P(W|x):=\delta_{D(x)}(w)$ rather than the prior $P(W)$, where $\delta(x)$ represents the Dirac delta function located at $x$. This forces the distribution of the features to match more precisely as compared to the simple first-moment matching proposed by Salimans et al. in \cite{DBLP:journals/corr/SalimansGZCRC16}.

\qheading{Addressing mismatch between prior and posterior:} Finally, we address the possibility of mismatch between inferred and prior latent distributions (point (3) described above), by imposing a maximum mean discrepancy (MMD) loss between the sets of samples of the said two distributions. We use an RBF kernel to compute this loss. This loss improves the random sampling quality by providing a direct learning signal to the mapping network.

Finally we summarize the objective of our complete model as in Equation \ref{eq:full_objective}. Here and in the rest of the paper we use a plus operator $+$, between two optimization process to indicate that both of them are performed simultaneously.
\begin{equation}
    \min_{G, M}\left[\max_{D}\mathcal{L}_\text{GAN} + \min_{D}\left[\mathbb{E}_{w \doubleplus w_R}\left[\norm{M(z) - \tilde{w}}_2^2 + \mathcal{L}_\mathrm{fm} + \norm{\tilde{w} - \tilde{\tilde{w}}}_2^2 + \mathrm{MMD}\{w, w_R\}\right]\right]\right]
\label{eq:full_objective}
\end{equation}

\qheading{An optimal transport based interpretation:} Neglecting the last three terms described in Equation \ref{eq:full_objective}, our method can be interpreted as a Wasserstein autoencoder-GAN (WAE-GAN) \cite{tolstikhin2018wasserstein}. Considering a WAE with its data domain set to our latent space and its latent space assigned to our image domain, if the encoder and the discriminator share weights the analogy is complete. Our model can, hence, be thought of as learning the latent variable model $P(W)$ by randomly sampling a data point $x\sim \sX$ from the training set and mapping it to a latent code $w$ via a deterministic transformation. In terms of density it can be written as in Equation \ref{eq:density_eq}.
\begin{equation}
    P(W) := \int_{x\in \sX} P(w|x)P(x)\mathrm{d}x. %
\label{eq:density_eq}
\end{equation}

As proven by Olivier, al. \cite{bousquet2017optimal}, under this model the optimal transport problem $W_c(P(W), P_D(W)) := \inf_{\Gamma\in P(w_1\sim P(W), w_2\sim P_D(W))}\left[\mathbb{E}_{w_1, w_2\sim\Gamma} \left[c(w_1, w_2)\right]\right]$ can be solved by finding a generative model $G(X|W)$ such that its $X$ marginal, $P_G(X) = \mathbb{E}_{w\sim P(W)}G(X|w)$ matches the image distribution $P(X)$. We ensure this by considering the Jensen–Shannon divergence $D_\text{JS}(P_G(X), P(X))$ using a GAN framework. This leads to the cost function given in Equation \ref{eq:ot_cost}, when we choose the ground cost function $c(w_1, w_2)$ to be squared $L_2$ norm.

\begin{equation}
    \min_{G, M}\max_{D}\mathcal{L}_\mathrm{GAN} + \min_{G, M}\min_{D}\norm{w - \tilde{w}}_2^2
\label{eq:ot_cost}
\end{equation}
Finally we find that by running the encoding/decoding cycle one more time we can impose several constraints that improve the quality of the encoder and the decoder network in practice. This leads to our full optimization criterion as described in Equation \ref{eq:full_objective}.

\subsection{Dealing with resolutions higher than training resolution}
Although StyleGAN \cite{karras2019analyzing} and BigGAN \cite{DBLP:journals/corr/abs-1809-11096} have shown that it is possible to generate relatively high resolution images, in the range of $1024 \times 1024$ and $512 \times 512$, their training is resource intense and the models are difficult to tune for new data sets. Equipped with invertability, we explore a tiling strategy to improve output resolution. First, we train an invertible GAN at a lower resolution ($m \times m$) and simply tile them $n \times n$ times with $n^2$ latent codes to obtain a higher resolution ($mn \times mn$) final output image. The new latent space containing $n^2$ latent codes obtained using the inference mechanism of the invertible GAN can now be used for various purposes as described in Section \ref{sec:tiled_latent_space} and reconstructions are visualized in Figure \ref{fig:recon_imgnet_ffhq_tiled}. This process correlates in spirit somewhat to COCO-GAN  \cite{DBLP:journals/corr/abs-1904-00284}. 
The main difference however is that our model at no point learns to assemble neighbouring patches. Indeed the seams are visible if one squints at the generated images, e.g in figure \ref{fig:recon_imgnet_ffhq_tiled}.
However, a detailed study of tiling for generation of higher resolution images than input domain is beyond the scope of our paper. We simply explore some naive settings and their applications in section \ref{sec:tiled_latent_space}.

\section{Experiments}
We test InvGAN on several diverse datasets (MNIST, ImageNet, CelebA, FFHQ) and multiple backbone architectures (DC-GAN, BigGAN, StyleGAN). Our method is evaluated both qualitatively (via style mixing, image inpainting etc.) and quantitatively (via the FID score and the suitability for data augmentation for discriminative tasks such as classification). Table \ref{tab:fids_and_recon_errors} shows random sample FIDs, middle point linear interpolation FIDs and test set reconstruction mean absolute errors (MAEs) of our generative model. 
We intend to provide a definition, baseline and understanding of inversion of a high quality generator. Specifically we highlight model-based inversion, joint training of generative and inference model and its usability in downstream tasks. We demonstrate that our method generalizes across architectures, datasets and types of downstream task.

\begin{table}[]
      \centering
      \begin{adjustbox}{max width=\textwidth}
        \addtolength{\leftskip} {-2cm}
        \addtolength{\rightskip}{-2cm}
        \begin{tabular}{l c c c c c c}
            \toprule
             \textbf{Models} & \textbf{RandFID}\newline & \textbf{RandRecFID} & \textbf{TsRecFID} & \textbf{IntTsFID} & \textbf{MAE$\pm 1$} & \textbf{Run Time}\\
             \midrule
             FFHQ \cite{xu2021generative} & 49.65/14.59 & 56.71/23.93 & -/13.73 & 68.45/38.01 & 0.129 & 0.045 \\
             FFHQ Enc.\cite{zhu2020indomain} & 46.82/14.38 & -/- & 88.48 / - & -/- & 0.460 \\
             FFHQ MSE opt.\cite{zhu2020indomain} & 46.82/14.38 & -/- & 58.04 / - & -/- & 0.106 \\
             FFHQ In-D. Inv.\cite{zhu2020indomain} & 46.82/14.38 & 52.02/- & 42.64 / - & 71.83/- & 0.115 & 99.76 \\
             
             \midrule
             DCGAN, MNIST & $17.44/6.10 $ & $16.76/4.25 $ & $17.77/4.70$ & $26.04/11.44 $  & $0.070$ & $3.3\cdot10^{-5}$\\
             StyleGAN, CelebA & $26.63/4.81 $ & $24.35/3.51 $ & $24.37/4.14 $ & $32.37/15.60 $ & $0.150$ & $1.0\cdot10^{-3}$\\ 
             StyleGAN, FFHQ & $49.14/12.12 $ & $44.42/8.85 $ & $41.14/7.15$ & $49.52/14.36 $ & $0.255 $ & $2.0\cdot10^{-3}$\\
             \bottomrule
        \end{tabular}
    \end{adjustbox}
        \caption{Here we report random sample FID (RandFID), FID of reconstructed random samples (RandRecFID), FID of reconstructed test set samples (TsRecFID), FID of the linear middle interpolation of test set images (IntTsFID) and reconstruction per pixel per color channel mean absolute error when images are normalized between $\pm 1$, also from test set. All FID scores are here evaluated against train set using 500 and 50000 samples. They are separated by `/'. For the traditional MSE optimization based and In-Domain GAN inversion, the MSE errors are converted to MAE by taking square root and averaging over the color channels and accounting for the re-normalization of pixel values between $\pm 1$. Runtime is given in seconds per image. We ran them on a V100 32GB GPU and measured wall clock time.}
    \label{tab:fids_and_recon_errors}
\end{table}

\qheading{Training data and tasks:}
We start with a StyleGAN-based architecture on FFHQ and CelebA for image editing. Then we train a BigGAN-based architecture on ImageNet, and show super resolution and video key-framing by tiling in the latent domain to work with images and videos that have higher resolution than training data. We also show ablation studies with a DC-GAN-based architecture on MNIST. 
This variety of architecture, dataset, and task provide insights into the method and its generality.
In the following sections we evaluate qualitatively by visualizing semantic editing of real images and quantitatively on various downstream tasks including classification fairness, image super resolution, image mixing, etc. %

\subsection{Semantically consistent inversion using InvGAN}
\label{sec:celeba_exp}
GANs can be used to augment training data and substantially improve downstream tasks learning.
Improving fairness of classifiers on human face images is a prominent example \cite{sharmanska2020contrastive,sattigeri2018fairness,balakrishnan2020towards,ramaswamy2020fair}. 
There is an important shortcoming in using existing GAN approaches for such tasks:
the labeling of augmented data relies on methods that are trained independently on the original data set, using human annotators or compute-expensive optimization-based inversion. This is due to the fact that most generative models used are unconditional and so generate unlabeled synthetic data.
A typical example is data-set de-biasing by Ramaswamy et al.~\cite{ramaswamy2020fair}. 
For each training image, an altered example that differs in some attribute (\eg age, hair color, ...) has to be generated. 
This has previously been done in one of two ways, \eg by finding the latent representation of the ground truth image via optimization or by labeling random samples using pre-trained classifiers on the biased data set. Optimization-based methods are slow and not a viable option for on-demand/adaptive data augmentation. Methods using pre-trained classifiers inherit their flaws like correlation induced dependencies.

Here we focus on the subproblem of reliably encoding face images to the latent space in a semantically consistent manner using InvGAN. For this we train ResNet50 attribute classifiers on the CelebA dataset. We validate that the encoding and decoding of InvGAN results in a semantically consistent reconstruction by training the classifier only on reconstructions of the full training set. As a baseline, we use the same classifier trained on the original CelebA. We produce two reconstructed training sets by using the tiling based inversion (trained on ImageNet) and by training InvGAN on CelebA (without tiling). For each attribute a separate classifier has been trained for $20$ epochs. The resulting mean average precisions are reported in Table \ref{tab:results_semantic_reconstruction}. We see that training on the reconstructions allows for very good domain transfer to real images, indicating that the reconstruction process maintained the semantics of the images.
\begin{table}[]
      \centering
        \begin{tabular}{l c c c}
            \toprule
             \textbf{Evaluated on} & \textbf{Original} & \textbf{Tile Reconstruction} & \textbf{Full Reconstruction} \\ \midrule
              \textbf{Original} & $0.81 \pm 0.15$ & $0.77 \pm 0.16$ & $0.79 \pm 0.15$ \\
            \textbf{Tile recon.} & $0.79 \pm 0.16$ & $0.80 \pm 0.15$ & $0.78 \pm 0.16$ \\
            \textbf{Full recon.} & $0.81 \pm 0.15$ & $0.78 \pm 0.16$ & $0.81 \pm 0.14$ \\ \bottomrule
            \textbf{Recon. Vis.}
            &
            \multicolumn{1}{m{3.2cm}}{
            \rule{0pt}{1.6cm}
            \includegraphics[width=1.5cm, bb= 0 0 128 128]{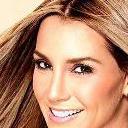} \includegraphics[width=1.5cm]{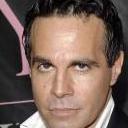}}
            &
            \multicolumn{1}{m{3.2cm}}{
            \rule{0pt}{1.6cm}
            \includegraphics[width=1.5cm]{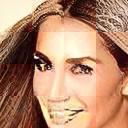} \includegraphics[width=1.5cm]{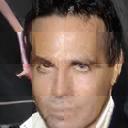}}
            &
            \multicolumn{1}{m{3.2cm}}{
            \rule{0pt}{1.6cm}
            \includegraphics[width=1.5cm]{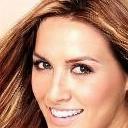} \includegraphics[width=1.5cm]{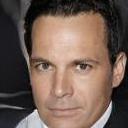}}
        \end{tabular}

        \caption{Mean average precision for a ResNet50 attribute classifier on CelebA, averaged over 20 attributes. We report the performance for training on the original dataset, the reconstructed dataset using the tiling based method pre-trained on ImageNet and the reconstruction on InvGAN trained on the CelebA training set directly.}
    \label{tab:results_semantic_reconstruction}
\end{table}

\subsection{Suitability for image editing}
GAN inversion methods have been proposed for machine supported photo editing tasks \cite{zhu2020indomain, cheng2020sequential, perarnau2016invertible}. Although there is hardly any quantitative evaluation for the suitability of a specific inversion algorithm or model, a variety of representative operations have been reported \cite{DBLP:journals/corr/abs-1911-11544, DBLP:journals/corr/abs-1904-03189, zhu2020indomain}. Amongst those are in-painting cut out regions of an image, image-merging and improving on amateurish manual photo editing. Figures \ref{fig:photo_editing_ffhq} and \ref{fig:photo_editing_celeba} in appendix visualize those operations performed on FFHQ and CelebaA images respectively. We demonstrate in-painting by zeroing out a randomly positioned square patch and then simply reconstructing the image. This can be interpreted as a image-repair operation/correcting imperfections in unseen data. The image-merging is performed by reconstructing an image which is composed out of two images by simply placing them together. By reconstructing an image that has undergone manual photo editing, higher degrees of photo-realism are achieved. Quantitative metric for such tasks are hard to define and hence is scarcely found in prior art, since they depend upon visual quality of the results. We report reconstruction and interpolation FIDs in Table \ref{tab:fids_and_recon_errors}, in an effort to establish a baseline for future research. However, wee do acknowledge that a boost in pixel fidelity in our reconstruction will greatly boost the performance of InvGAN on photo editing tasks. The experiments clearly show the general suitability of the learned representations to project out of distribution images to the learned posterior manifold via reconstruction.

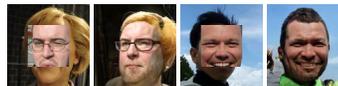
\begin{figure}[h!]
\begin{center}
\begin{adjustbox}{max width=1\linewidth}
  \begin{tikzpicture}[
    thick, text centered,
    box/.style={draw, minimum width=0.7cm, minimum height=0.7cm},
    box_image/.style={draw, minimum width=1.1cm, minimum height=1.1cm},
    func/.style={circle, text=white},
  ]
    
    \node (x00_title) at (0,0) [text width=2.4cm, align=left] {\textbf{Style mixing}};

	\node (x10_title) [text width=2.4cm, below=1.65cm of x00_title, align=left] {\textbf{In-painting}};
	\node (x10) [right=-0.2cm of x10_title] {\includegraphics[width=1.2cm]{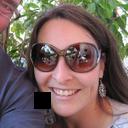}};
	\node (x11) [right=-0.2cm of x10] {\includegraphics[width=1.2cm]{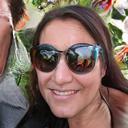}};
	\node (x12) [right=-0.2cm of x11] {\includegraphics[width=1.2cm]{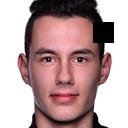}};
	\node (x13) [right=-0.2cm of x12] {\includegraphics[width=1.2cm]{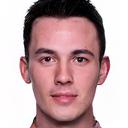}};

    \node (x30_title) [text width=2.26cm, right=0.cm of x13, align=left] {\textbf{Editing}};
	\node (x30) [right=-0.2cm of x30_title] {\includegraphics[width=1.2cm]{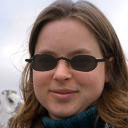}};
	\node (x31) [right=-0.2cm of x30] {\includegraphics[width=1.2cm]{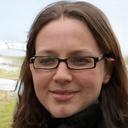}};
	\node (x32) [right=-0.2cm of x31] {\includegraphics[width=1.2cm]{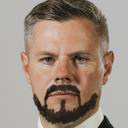}};
	\node (x33) [right=-0.2cm of x32] {\includegraphics[width=1.2cm]{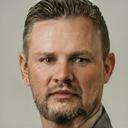}};

	\node (x20_title) [text width=2.4cm, below=0.8cm of x10_title, align=left] {\textbf{Merging I}};
	\node (x20) [right=-0.2cm of x20_title] {\includegraphics[width=1.2cm]{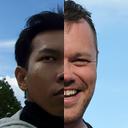}};
	\node (x21) [right=-0.2cm of x20] {\includegraphics[width=1.2cm]{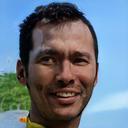}};
	\node (x22) [right=-0.2cm of x21] {\includegraphics[width=1.2cm]{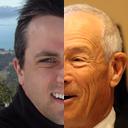}};
	\node (x23) [right=-0.2cm of x22] {\includegraphics[width=1.2cm]{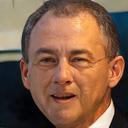}};
	
	\node (x20_title2) [text width=2.26cm, right=0.cm of x23, align=left] {\textbf{Merging II}};
	\node (x24) [right=-0.2cm of x20_title2] {\includegraphics[width=1.2cm]{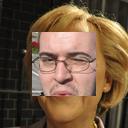}};
	\node (x25) [right=-0.2cm of x24] {\includegraphics[width=1.2cm]{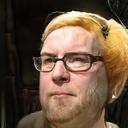}};
	\node (x26) [right=-0.2cm of x25] {\includegraphics[width=1.2cm]{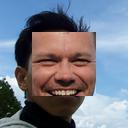}};
	\node (x27) [right=-0.2cm of x26] {\includegraphics[width=1.2cm]{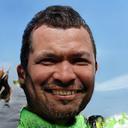}};

	\node (x02) [above=-0.1cm of x10] {\includegraphics[width=1.2cm]{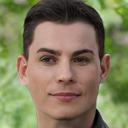}};
	\node (x01) [left=-0.2cm of x02] {\includegraphics[width=1.2cm]{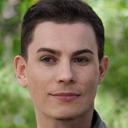}};	
	\node (x00) [left=-0.2cm of x01] {\includegraphics[width=1.2cm]{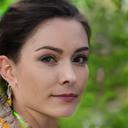}};
	\node (x03) [right=-0.2cm of x02] {\includegraphics[width=1.2cm]{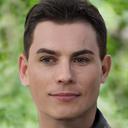}};
	\node (x04) [right=-0.2cm of x03] {\includegraphics[width=1.2cm]{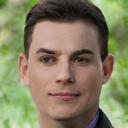}};
	\node (x05) [right=-0.2cm of x04] {\includegraphics[width=1.2cm]{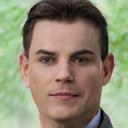}};
	\node (x06) [right=-0.2cm of x05] {\includegraphics[width=1.2cm]{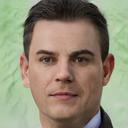}};
	\node (x07) [right=-0.2cm of x06] {\includegraphics[width=1.2cm]{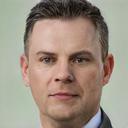}};
	\node (x08) [right=-0.2cm of x07] {\includegraphics[width=1.2cm]{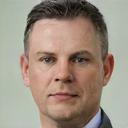}};
	\node (x09) [right=-0.2cm of x08] {\includegraphics[width=1.2cm]{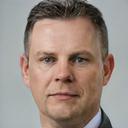}};
	\node (x010) [right=-0.2cm of x09] {\includegraphics[width=1.2cm]{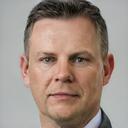}};
	\node (x011) [right=-0.2cm of x010] {\includegraphics[width=1.2cm]{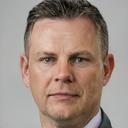}};

\end{tikzpicture}	
\end{adjustbox}
\end{center}
	\caption{Benchmark image editing tasks on FFHQ ($128\,\mathrm{px}$). Style mixing: We transfer the first $0, 1, 2, \dots, 11$ style vectors from one image to another. For the other image editing tasks, pairs of images are input image (left) and reconstruction (right).}
	\label{fig:photo_editing_ffhq}
\end{figure}

\subsection{Tiling to boost resolution}
\label{sec:tiled_latent_space}
Limitations in video RAM and instability of high resolution GANs are prominent obstacles in generative model training. One way to bypass such difficulties is to generate the image in parts. Here we train our invertible generative model, a BigGAN architecture on $32\times 32$ random patches from ImageNet. Once the inversion mechanism and the generator are trained to satisfactory quality, we reconstruct both FFHQ and ImageNet images. We use $256 \times 256$ resolution and tiling $64$ patches in an $8\times 8$ grid for FFHQ images, and $128\times 128$ resolution and tiling $16$ patches in a $4\times 4$ grid for ImageNet images. The reconstruction results are shown in Figure \ref{fig:recon_imgnet_ffhq_tiled}.
Given the successful reconstruction process, we explore the tiled latent space for tasks such as image deblurring and time interpolation of video frames.

\begin{figure*}
    \centering
    \includegraphics[width=\linewidth]{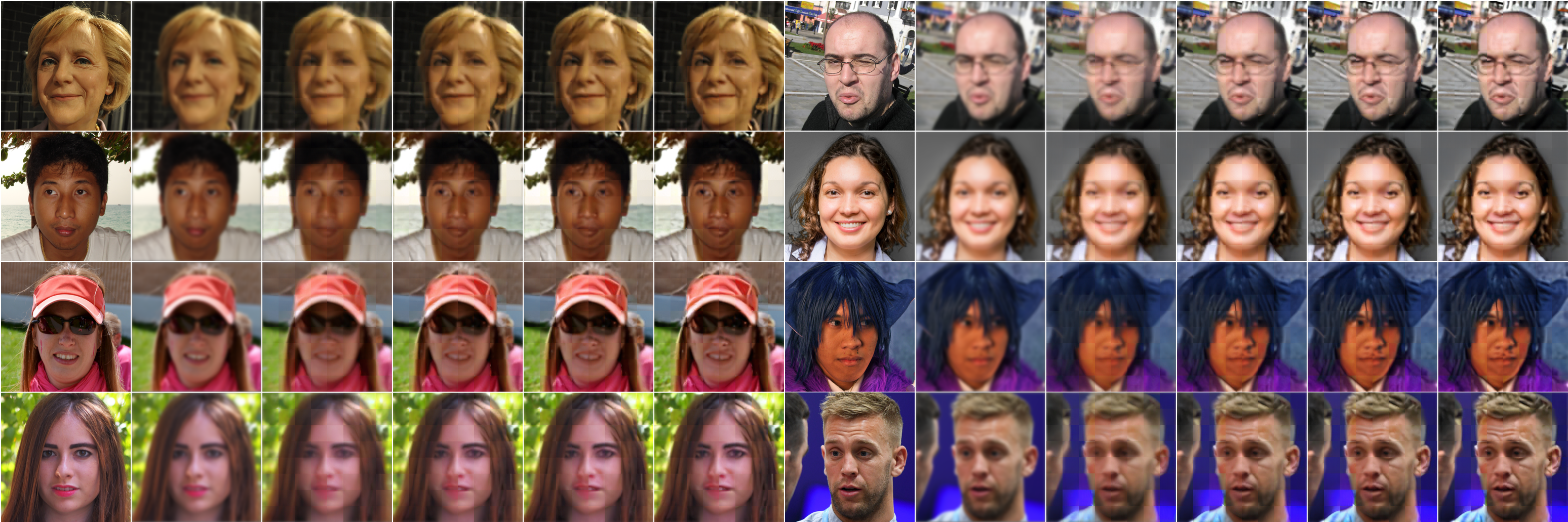}
    \caption{Super resolution using extrapolation in the tiled latent space. From left we visualize the original image, the low resolution version of it, the reconstruction of the low resolution version, and progressive extrapolation to achieve debluring.}
    \label{fig:super_resolution}
\end{figure*}

\qheading{Image de-blurring:}
Here we take a low resolution image, scale it to the intended resolution using bicubic interpolation, invert it patch by patch, gaussian blur it, invert it again and linearly extrapolate it in the deblurring direction. The deblurring direction is simply obtained by subtracting the latent code of the given low resolution but bicubic up sampled image from the latent code of the blurred version of it at the same resolution. The exact amount of extrapolation desired is left up to the user. As shown in Figure \ref{fig:super_resolution} we show the effect of $3$ different levels of extrapolation. Although our method is not trained for the task of super resolution, by virtue of a meaningful latent space we can enhance image quality.

\qheading{Temporal interpolation of video frames:}
Here we boost the frame rate of a video post capture. We infer the tiled latent space of consecutive frames in a video and linearly interpolate each tile to generate one or more intermediate frames. Results are shown in the accompanying videos in the supplementary material and in Figure \ref{fig:video_time_resmpl} in the appendix. This can be used to create slow motion video, post capture. We find the latent code of each frame in a video sequence and then derive intermediate latent codes by weighted averaging neighbouring latent codes using a Gaussian window. Even though this effectively interpolates between latent codes for background patch with foreground patch for fast moving small objects leading to blur, it results in smooth slow motion video, as can be seen in the supplementary material . We use the UCF101 data set \cite{DBLP:journals/corr/abs-1212-0402} for this task.

\begin{figure*}
    \centering
      \begin{subfigure}{0.5\linewidth}
        \includegraphics[width=\linewidth]{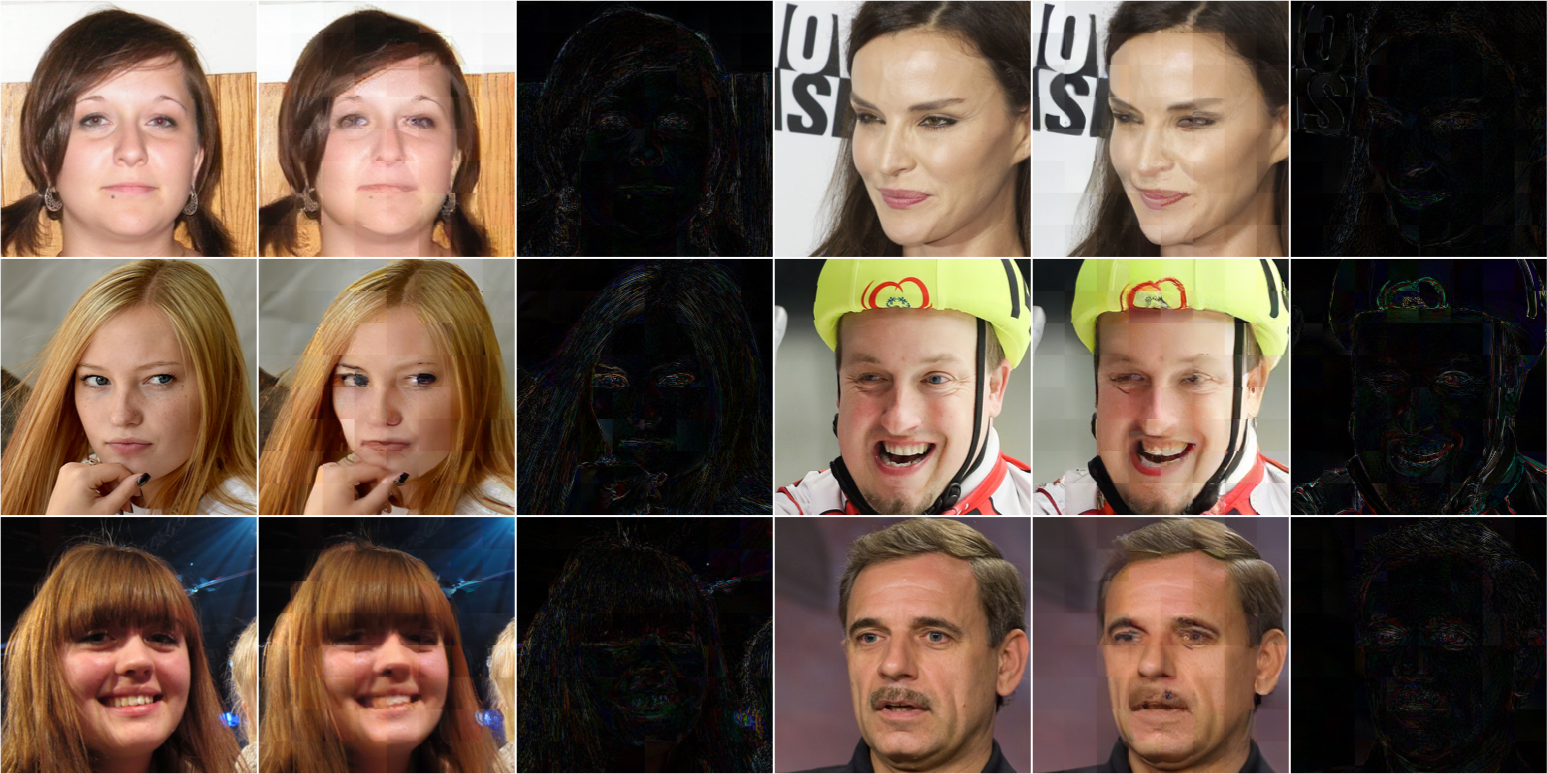}
        \caption{}
        \label{fig:sub_ffhq}
      \end{subfigure}%
      ~
      \begin{subfigure}{0.46\linewidth}
        \includegraphics[width=\linewidth]{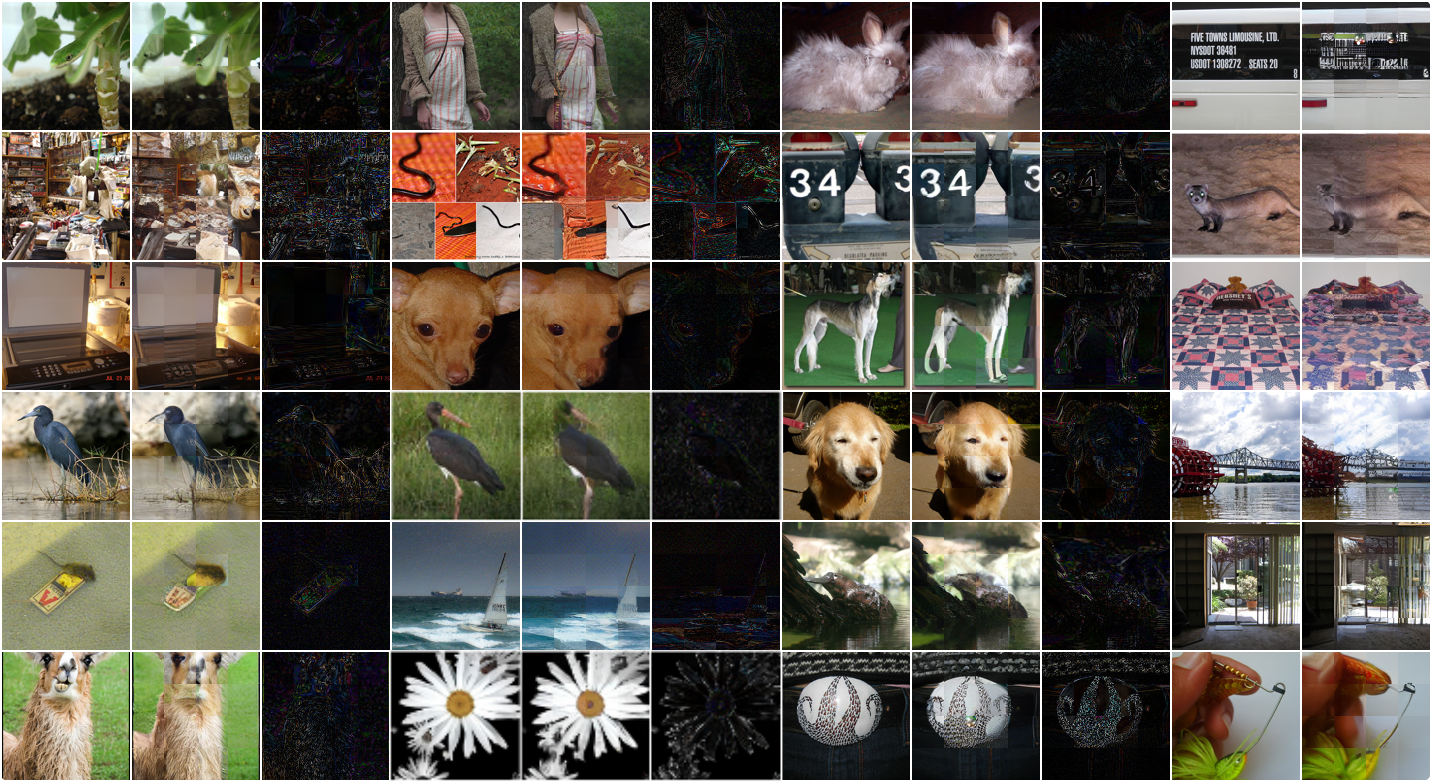}
        \caption{}
        \label{fig:sub_ffhq}
      \end{subfigure}%
      
    \caption{Tiled reconstruction of random (a) FFHQ Images and (b) ImageNet images. Left column shows the real images, the second shows the patch by patch reconstructions and the third shows the absolute pixel-wise differences. Note that interestingly though the patches are reconstructed independent of each other, the errors lie mostly on the edges of the objects in the images, arguably the most information dense region of the images.}
    \label{fig:recon_imgnet_ffhq_tiled}
\end{figure*}

\subsection{Ablation studies}
\label{sec:ablation_resue_real_z}
\begin{figure*}
    \centering
      \begin{subfigure}{0.33\linewidth}
        \includegraphics[width=\linewidth]{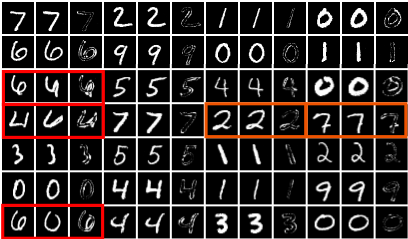}
        \caption{}
        \label{fig:sub_mnist_a}
      \end{subfigure}%
      ~
      \begin{subfigure}{0.33\linewidth}
        \includegraphics[width=\linewidth]{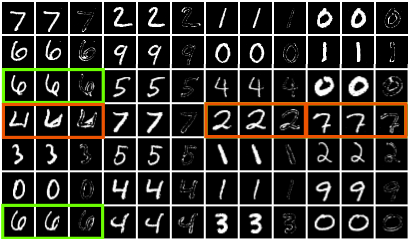}
        \caption{}
        \label{fig:sub_ffhq_b}
      \end{subfigure}%
      ~
      \begin{subfigure}{0.33\linewidth}
        \includegraphics[width=\linewidth]{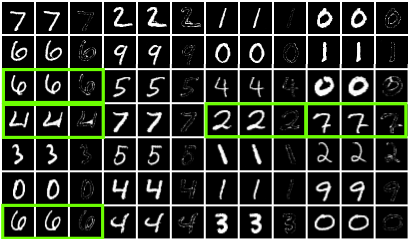}
        \caption{}
        \label{fig:sub_ffhq_c}
      \end{subfigure}%
      
    \caption{Inversion of held out test samples. Columns are in groups of three: first column holds real images, second their reconstruction and third the absolute pixel-wise difference. (a) Inversion using naive model, i.e. only z reconstruction loss is used, (b) inversion using model that uses latent codes from real samples, i.e. the augmented naive model. (c) our full model. Notice how the imperfections in the reconstructions highlighted with red boxes gradually vanishes as the model improves.}
    \label{fig:recon_imgnet_ffhq_tiled}
\end{figure*}

Recall that the naive model defined in Section \ref{sub_sec:objective} uses the following optimization $\min_{G,M}\left[\max_D L_\text{GAN} + \min_D\mathbb{E}_{z \sim P(Z)}\norm{M(z) - \tilde{w}}_p\right]$ to train (also given in Equation \ref{eq:ot_cost}), (results in Figure \ref{fig:sub_mnist_a}). Here we progressively show how our three main components contributes on the naive model. As is apparent from the method section, the first major improvement comes from exposing the generator to the latent code inferred from real images. This is primarily due to the difference in the prior and the induced posterior distribution. This is especially true during early training, which imparts a lasting impact. The corresponding optimization is $\min_{G,M}\left[\max_D L_\text{GAN} + \min_D\mathbb{E}_{w=M(z \sim P(Z)) \doubleplus w_R}\norm{w - \tilde{w}}_p\right]$. Simply reducing the distribution mismatch between prior and posterior by injecting inferred latent codes improves inversion quality. This is visualized in Figure \ref{fig:sub_ffhq_b}. We shall call this model as the augmented naive model. However, this modification unlocks the possibility to enforce back propagation of generator loss gradients to the discriminator and real-image, generated-image pairing as detailed in Section \ref{sub_sec:objective}. This leads to our full model and the results are visualized in Figure \ref{fig:sub_ffhq_c}

\section{Discussion and future work}
While InvGAN can reliably invert the generator of a GAN, it still can benefit from an improved reconstruction fidelity for tasks such as, image compression, image segmentation, etc. We observe that the reconstruction of rare features, such as microphones, hats or background features, tend to have lower quality, as seen in appendix in Figure \ref{fig:recons_celebA} bottom row 3rd and 4th column. This combined with the fact that reconstruction loss during training tends to saturate even when the weights are sufficiently high indicates that even well-engineered architectures such as StyleGAN and BigGAN lack representative power to provide sufficient data coverage.

Strong inductive biases in the generative model have the potential to improve the quality of the inference module. For instance GIF~\cite{GIF2020} and hologan~\cite{nguyen2019hologan} among others introduce strong inductive bias from the underlying 3D geometry and lighting properties of a 2D image. Hence an inverse module of these generative mechanism has the potential to outperform their counterparts, which are trained fully supervised on the labelled training data alone at estimating 3D face parameters from 2D images.

As was shown by the success of RAEs~\cite{Ghoshetal19}, there is often a mismatch between the induced posterior and the prior of generative models which can be removed by an ex-post density estimator. InvGAN is also aminable to ex-post density estimation. When applied to the tiled latent codes, it estimates a joint density of the tiles for unseen data. This would recover a generative model without going through the unstable GAN training.

We have shown that our method scales to large datasets such as ImageNet, CelebA, and FFHQ. A future work that is able to improve upon reconstruction fidelity, would be able explore adversarial robustness by extending \cite{Ghosh_Losalka_Black_2019} to larger datasets.

\section{Conclusion}
We presented InvGAN, an inference framework for the latent generative parameters used by a GAN generator. InvGAN enjoys several advantages compared to state-of-the-art inversion mechanisms. The inversion mechanism is integrated into the training phase of the generator, potentially contributing to the disentanglement of the latent space. Beyond the computational advantage of model-based inversion, our mechanism can reconstruct images that are larger than the training images by tiling with no additional merging needed. We further demonstrated that the inferred latent code for a given image is semantically meaningful i.e. it falls inside the structured part of the latent space learned by the generator.

\begin{ack}
We thank Alex Vorobiov, Javier Romero, Betty Mohler Tesch and Soubhik Sanyal for their insightful comments and intriguing discussions. This research was fully funded and supported by Amazon. While PG and DZ are affiliated with Max Planck Institute for Intelligent Systems, this project was completed during PG's and DZ's internship at Amazon. MJB is affiliated to both Amazon and MPI. This work however was carried out solely with Amazon's support.
\end{ack}

\newpage
{\small
\bibliographystyle{ieee}
\bibliography{inv_gan_bibliography}
}

\clearpage
\appendix
\section*{Supplementary Material (InvGAN: Invertible GANs)}

\subsection*{Video Key Framing}

\begin{figure*}[h!]
    \centering
    \includegraphics[width=\linewidth]{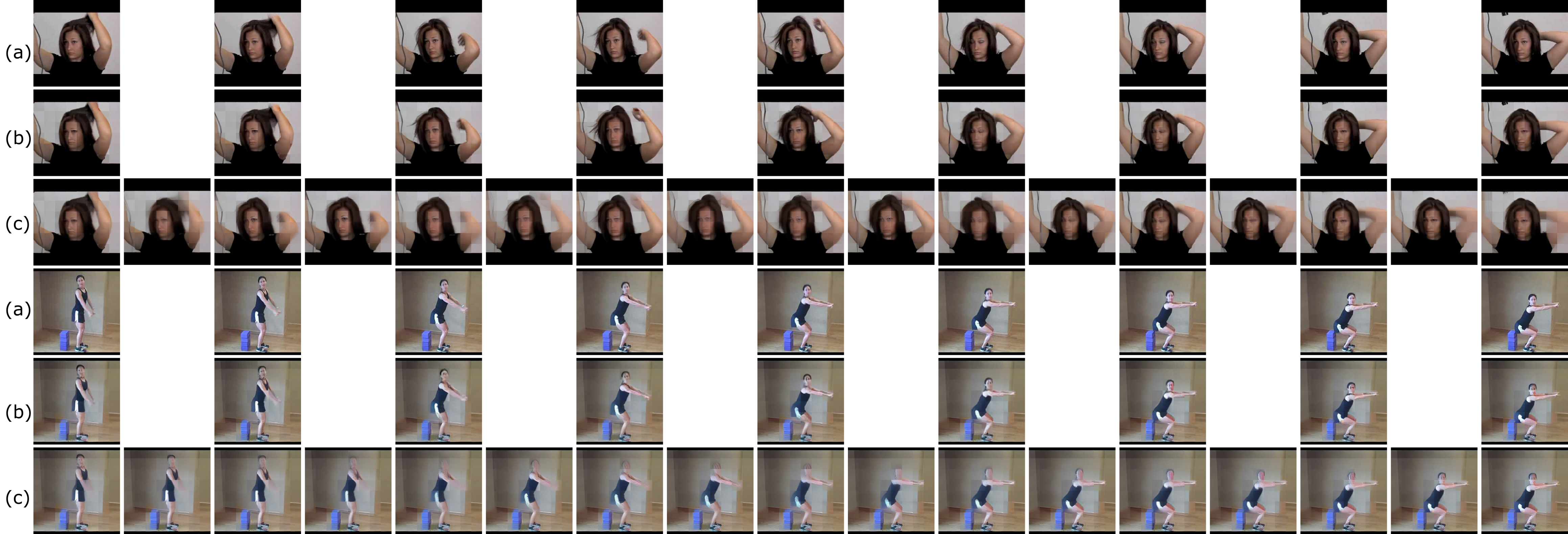}
    \caption{Tiled reconstruction of a video sequence. (a) original sequence, (b) reconstructed sequence, (c) up-sampled in time sequence.}
    \label{fig:video_time_resmpl}
\end{figure*}

\subsection*{CelebA Reconstructions}
\begin{center}
\addtolength{\tabcolsep}{-5pt}
\begin{tabular}{cccccccc}
 \includegraphics[width=0.12\linewidth]{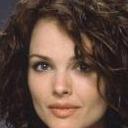} & \includegraphics[width=0.12\linewidth]{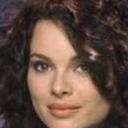} & \includegraphics[width=0.12\linewidth]{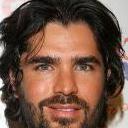} & \includegraphics[width=0.12\linewidth]{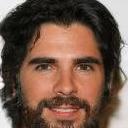} & \includegraphics[width=0.12\linewidth]{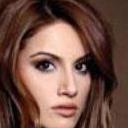} & \includegraphics[width=0.12\linewidth]{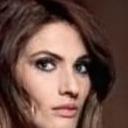} & \includegraphics[width=0.12\linewidth]{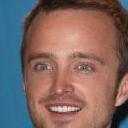} & \includegraphics[width=0.12\linewidth]{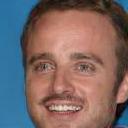} \\ 
 \includegraphics[width=0.12\linewidth]{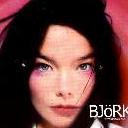} & \includegraphics[width=0.12\linewidth]{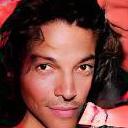} & \includegraphics[width=0.12\linewidth]{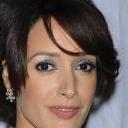} & \includegraphics[width=0.12\linewidth]{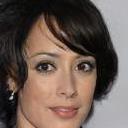} & \includegraphics[width=0.12\linewidth]{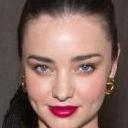} & \includegraphics[width=0.12\linewidth]{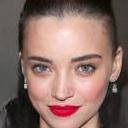} & \includegraphics[width=0.12\linewidth]{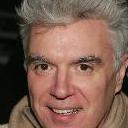} & \includegraphics[width=0.12\linewidth]{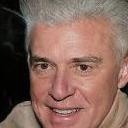} \\ 
  \includegraphics[width=0.12\linewidth]{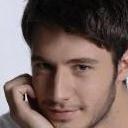} & \includegraphics[width=0.12\linewidth]{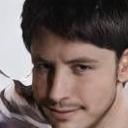} & \includegraphics[width=0.12\linewidth]{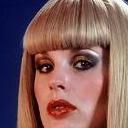} & \includegraphics[width=0.12\linewidth]{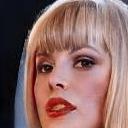} & \includegraphics[width=0.12\linewidth]{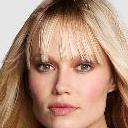} & \includegraphics[width=0.12\linewidth]{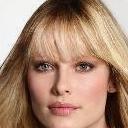} & \includegraphics[width=0.12\linewidth]{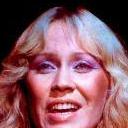} & \includegraphics[width=0.12\linewidth]{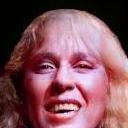} \\ 
   \includegraphics[width=0.12\linewidth]{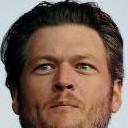} & \includegraphics[width=0.12\linewidth]{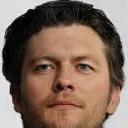} & \includegraphics[width=0.12\linewidth]{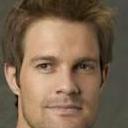} & \includegraphics[width=0.12\linewidth]{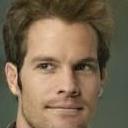} & \includegraphics[width=0.12\linewidth]{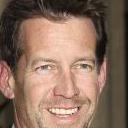} & \includegraphics[width=0.12\linewidth]{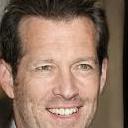} & \includegraphics[width=0.12\linewidth]{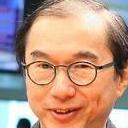} & \includegraphics[width=0.12\linewidth]{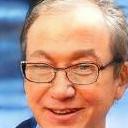} \\ 
    \includegraphics[width=0.12\linewidth]{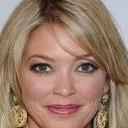} & \includegraphics[width=0.12\linewidth]{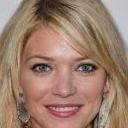} & \includegraphics[width=0.12\linewidth]{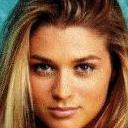} & \includegraphics[width=0.12\linewidth]{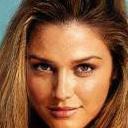} & \includegraphics[width=0.12\linewidth]{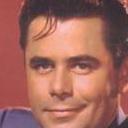} & \includegraphics[width=0.12\linewidth]{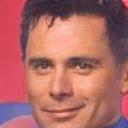} & \includegraphics[width=0.12\linewidth]{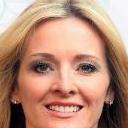} & \includegraphics[width=0.12\linewidth]{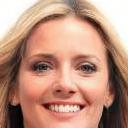} \\ 
     \includegraphics[width=0.12\linewidth]{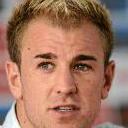} & \includegraphics[width=0.12\linewidth]{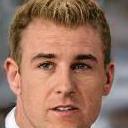} & \includegraphics[width=0.12\linewidth]{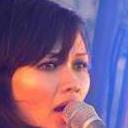} & \includegraphics[width=0.12\linewidth]{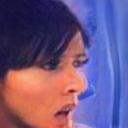} & \includegraphics[width=0.12\linewidth]{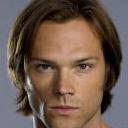} & \includegraphics[width=0.12\linewidth]{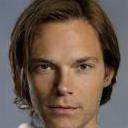} & \includegraphics[width=0.12\linewidth]{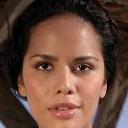} & \includegraphics[width=0.12\linewidth]{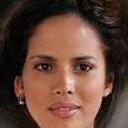} \\ 
\end{tabular}
\captionof{figure}{InvGAN reconstructions of CelebA at $128 \times 128$ resolution. Alternating (from left to right) original and reconstructed images.}
\label{fig:recons_celebA}
\end{center}
\newpage

\subsection*{Additional FID Evaluations}
We additionally evaluate the FID scores on reconstructions using the approach of ReStyle \cite{alaluf2021restyle}. The resulting scores are presented in Table \ref{tab:fids_and_recon_errors_extra}. Unfortunately we can not compute test set MAE and FIDs since ReStyle has been trained on the whole FFHQ set.
\begin{table}[h!]
      \centering
        \begin{tabular}{l c c c c c}
            \toprule
             \textbf{models} & \textbf{RandFID}\newline & \textbf{RandRecFID}\\
             \midrule
             1itr FFHQ ReStyle\cite{alaluf2021restyle} & 40.33/4.71 & 57.63/29.56 \\
             2itr FFHQ ReStyle\cite{alaluf2021restyle} & 40.33/4.71 & 53.09/22.88 \\
             3itr FFHQ ReStyle\cite{alaluf2021restyle} & 40.33/4.71 & 51.68/20.80 \\
             4itr FFHQ ReStyle\cite{alaluf2021restyle} & 40.33/4.71 & 51.49/19.93 \\
             \bottomrule
        \end{tabular}
        \caption{Random sample FID (RandFID), FID of reconstructed random samples (RandRecFID). FID scores are here evaluated using 500 and 50000 samples. They are separated by `/'.}
    \label{tab:fids_and_recon_errors_extra}
\end{table}

\subsection*{Additional Baseline for Semantic Reconstruction}
We conduct the same experiment as presented in Section \ref{sec:celeba_exp} with reconstructions using ReStyle \cite{alaluf2021restyle}. The resulting mean average precision evaluated on the reconstructed evaluation set is $0.80 \pm 0.15$. Evaluated on original evaluation set images, the performance drops to $0.78 \pm 0.17$, which indicates a weaker transfer as compared to both the tiled reconstruction and the full reconstruction using InvGAN.

\subsection*{CelebA Image Editing}
\begin{figure}[h!]
\begin{center}
\begin{adjustbox}{max width=1\linewidth}
  \begin{tikzpicture}[
    thick, text centered,
    box/.style={draw, minimum width=0.7cm, minimum height=0.7cm},
    box_image/.style={draw, minimum width=1.1cm, minimum height=1.1cm},
    func/.style={circle, text=white},
  ]
    
    \node (x00_title) at (0,0) [text width=2.4cm, align=left] {\textbf{Style mixing}};

	\node (x10_title) [text width=2.4cm, below=1.65cm of x00_title, align=left] {\textbf{In-painting}};
	\node (x10) [right=-0.2cm of x10_title] {\includegraphics[width=1.2cm]{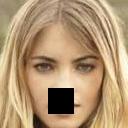}};
	\node (x11) [right=-0.2cm of x10] {\includegraphics[width=1.2cm]{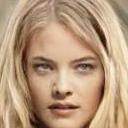}};
	\node (x12) [right=-0.2cm of x11] {\includegraphics[width=1.2cm]{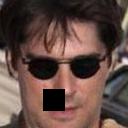}};
	\node (x13) [right=-0.2cm of x12] {\includegraphics[width=1.2cm]{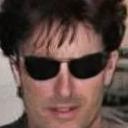}};

    \node (x30_title) [text width=2.26cm, right=0.cm of x13, align=left] {\textbf{Editing}};
	\node (x30) [right=-0.2cm of x30_title] {\includegraphics[width=1.2cm]{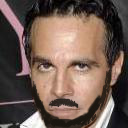}};
	\node (x31) [right=-0.2cm of x30] {\includegraphics[width=1.2cm]{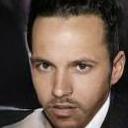}};
	\node (x32) [right=-0.2cm of x31] {\includegraphics[width=1.2cm]{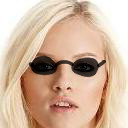}};
	\node (x33) [right=-0.2cm of x32] {\includegraphics[width=1.2cm]{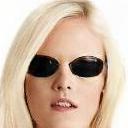}};

	\node (x20_title) [text width=2.4cm, below=0.8cm of x10_title, align=left] {\textbf{Merging I}};
	\node (x20) [right=-0.2cm of x20_title] {\includegraphics[width=1.2cm]{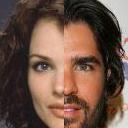}};
	\node (x21) [right=-0.2cm of x20] {\includegraphics[width=1.2cm]{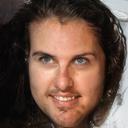}};
	\node (x22) [right=-0.2cm of x21] {\includegraphics[width=1.2cm]{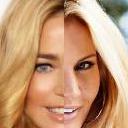}};
	\node (x23) [right=-0.2cm of x22] {\includegraphics[width=1.2cm]{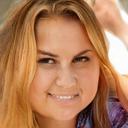}};
	
	\node (x20_title2) [text width=2.26cm, right=0.cm of x23, align=left] {\textbf{Merging II}};
	\node (x24) [right=-0.2cm of x20_title2] {\includegraphics[width=1.2cm]{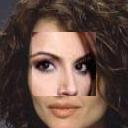}};
	\node (x25) [right=-0.2cm of x24] {\includegraphics[width=1.2cm]{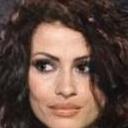}};
	\node (x26) [right=-0.2cm of x25] {\includegraphics[width=1.2cm]{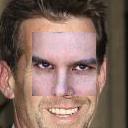}};
	\node (x27) [right=-0.2cm of x26] {\includegraphics[width=1.2cm]{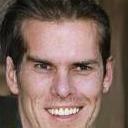}};

	\node (x02) [above=-0.1cm of x10] {\includegraphics[width=1.2cm]{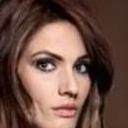}};
	\node (x01) [left=-0.2cm of x02] {\includegraphics[width=1.2cm]{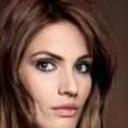}};	
	\node (x00) [left=-0.2cm of x01] {\includegraphics[width=1.2cm]{figures/celeba_appendix/stylemix/49_upto_0_stylemixed.jpg}};
	\node (x03) [right=-0.2cm of x02] {\includegraphics[width=1.2cm]{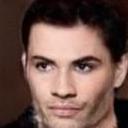}};
	\node (x04) [right=-0.2cm of x03] {\includegraphics[width=1.2cm]{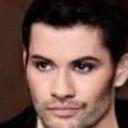}};
	\node (x05) [right=-0.2cm of x04] {\includegraphics[width=1.2cm]{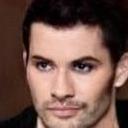}};
	\node (x06) [right=-0.2cm of x05] {\includegraphics[width=1.2cm]{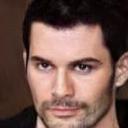}};
	\node (x07) [right=-0.2cm of x06] {\includegraphics[width=1.2cm]{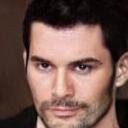}};
	\node (x08) [right=-0.2cm of x07] {\includegraphics[width=1.2cm]{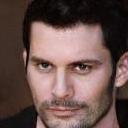}};
	\node (x09) [right=-0.2cm of x08] {\includegraphics[width=1.2cm]{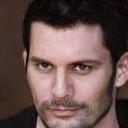}};
	\node (x010) [right=-0.2cm of x09] {\includegraphics[width=1.2cm]{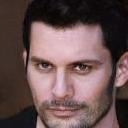}};
	\node (x011) [right=-0.2cm of x010] {\includegraphics[width=1.2cm]{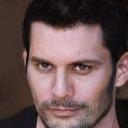}};

\end{tikzpicture}	
\end{adjustbox}
\end{center}
	\caption{Benchmark image editing tasks on CelebA ($128\,\mathrm{px}$). Style mixing: We transfer the first $0, 1, 2, \dots, 11$ style vectors from one image to another. For the other image editing tasks, pairs of images are input image (left) and reconstruction (right).}
	\label{fig:photo_editing_celeba}
\end{figure}

We conducted the same image editing operations shown in Figure \ref{fig:photo_editing_ffhq} on CelebA. The results are shown in Figure \ref{fig:photo_editing_celeba}.

\end{document}